\documentclass[runningheads]{llncs}
\usepackage[T1]{fontenc}
\usepackage{multirow}
\usepackage{graphicx}
\usepackage{amsmath}
\usepackage{enumitem}
\usepackage{url}
\usepackage{tabularx}
\usepackage{ragged2e}
\usepackage{subcaption}
\usepackage{amsfonts}
\usepackage{booktabs}
\usepackage{amssymb}
\usepackage{pifont}
\begin{document}
\title{Learning Granularity Representation for Temporal Knowledge Graph Completion}
\titlerunning{LGRe}
%
\author{Jinchuan Zhang\inst{1} \and
Tianqi Wan\inst{1,2} \and
Chong Mu\inst{3} \and
Guangxi Lu\inst{1} \and
Ling Tian\inst{1,4} }
\authorrunning{Jinchuan Zhang et al.}
%
\institute{School of Computer Science and Engineering, University of Electronic Science and Technology of China, Chengdu, China \and
Beijing Institute of Aerospace Measurement and Testing Technology, Beijing, China
\and School of Information and Software Engineering, University of Electronic Science and Technology of China, Chengdu, China 
\and KASH Institute of Electronics and Information Industry, Kashi, China
\\
\email{\{jinchuanz,muchong\}@std.uestc.edu.cn}\\
\email{\{18600016300\}@163.com}\\
\email{\{glu3,lingtian\}@uestc.edu.cn}}
\maketitle              
\begin{abstract}
Temporal Knowledge Graphs (TKGs) incorporate temporal information to reflect the dynamic structural knowledge and evolutionary patterns of real-world facts. Nevertheless, TKGs are still limited in downstream applications due to the problem of incompleteness. Consequently, TKG completion (also known as link prediction) has been widely studied, with recent research focusing on incorporating independent embeddings of time or combining them with entities and relations to form temporal representations. However, most existing methods overlook the impact of history from a multi-granularity aspect. The inherent semantics of human-defined temporal granularities, such as ordinal dates, reveal general patterns to which facts typically adhere. To counter this limitation, this paper proposes \textbf{L}earning \textbf{G}ranularity \textbf{Re}presentation (termed $\mathsf{LGRe}$) for TKG completion. It comprises two main components: Granularity Representation Learning (GRL) and Adaptive Granularity Balancing (AGB). Specifically, GRL employs time-specific multi-layer convolutional neural networks to capture interactions between entities and relations at different granularities. After that, AGB generates adaptive weights for these embeddings according to temporal semantics, resulting in expressive representations of predictions. Moreover, to reflect similar semantics of adjacent timestamps, a temporal loss function is introduced. Extensive experimental results on four event benchmarks demonstrate the effectiveness of $\mathsf{LGRe}$ in learning time-related representations.
To ensure reproducibility, our code is available at https://github.com/KcAcoZhang/LGRe.

\keywords{Temporal Knowledge Graph \and Knowledge Graph Completion \and Representation Learning \and Link Prediction.}
\end{abstract}
\section{Introduction}
Temporal Knowledge Graphs (TKGs) \cite{kgsurvey,tkgsurvey} extend KGs with temporal information $t$, to indicate the dynamics and evolutionary patterns of real-world facts. TKGs commonly represent facts as quadruples like (\textit{subject}, \textit{relation}, \textit{object}, \textit{timestamp}), i.e., $(s,r,o,t)$, broadening the perspective of knowledge-based downstream applications \cite{llmtkgr,flkgc}. Nevertheless, TKGs still face the challenge of incompleteness, necessitating efficient methods to complete missing quadruples. Involving reasoning about missing entities at known or future timestamps, the link prediction task of TKGs can be divided into interpolation and extrapolation. Notably, this paper focuses on interpolation, also known as TKG completion, which can be regarded as a temporal extension of KG completion.

Addressing the incompleteness of TKGs necessitates consideration of the temporal evolution of facts. Consequently, numerous methods conceptualize dynamics as transformations within semantic space, incorporating temporal information into the translation \cite{ttranse,tadistmult}, rotation \cite{chronor,tero,telm}, or decomposition \cite{bdme,jmt,cec} of facts in Euclidean or Complex spaces, employing specialized score functions. Building upon these KG-derived approaches, more advanced techniques utilize Convolutional Neural Networks (CNNs) \cite{SANe} or Graph Neural Networks (GNNs) \cite{temp} to uncover more complex patterns within temporal information. However, most existing approaches treat time as an independent embedding, merely considering its fusion with entities and relations to represent dynamic knowledge. Furthermore, they often overlook the latent impact of multi-granularity temporal information (e.g., ordinal dates) inherent in real-world facts.

\begin{figure*}[b!]
\centerline{\includegraphics[width=0.6\linewidth]{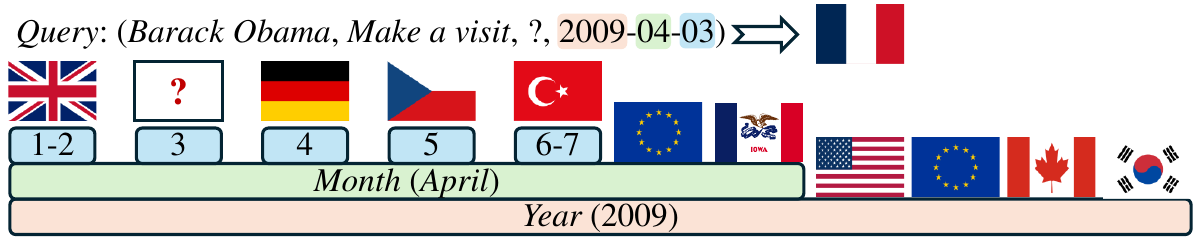}}
\caption{An illustration of TKG completion on ICEWS05-15, where the displayed flags indicate the known objects from the three-level granularity history.}
\label{example}
\end{figure*}
In reality, the periodicity and variability in these granularities reveal general patterns to which facts typically adhere, with the semantics of real-world facts being closely associated with their multi-granularity temporal attributes. To make accurate predictions, these granularities should be considered collectively. For instance, as illustrated in Figure \ref{example}, the query (\textit{Barack Obama}, \textit{Make a visit}, ?) exhibits varying semantics at different granularities (\textit{year}, \textit{month}, and \textit{day}) of history. Based on the learned historical knowledge, at the \textit{year} level, potential objects could include \textit{US}, \textit{European Union}, or \textit{Canada}, among others. Subsequently, the \textit{month}-level information might narrow the object down to the first two. At the \textit{day} level, while adjacent timestamps exhibit similar semantics, the ground truth might be a country geographically proximate to \textit{UK} and \textit{Germany}. Although the finest granularity provides the most relevant information, coarser granularities still contribute significant knowledge, such as the \textit{European Union} at the \textit{year} and \textit{month} levels. Consequently, we suggest that representations from each granularity may embody distinct meanings, all of which contribute to the prediction. Overlooking these granularities could potentially lead to neglecting important differences or similarities in representations of certain facts.

To tackle these limitations, we propose an effective method that delves into \textbf{L}earning \textbf{G}ranularity \textbf{Re}presentation ($\mathsf{LGRe}$). The model consists of two consecutive modules centered around temporal information: a) Granularity Representation Learning (GRL) that employs multi-layer CNNs parametered by temporal semantics, to capture the correlations among entities, relations and time within facts and further learn representations belonging to different granularities (i.e., \textit{year}, \textit{month} or \textit{day}); b) Adaptive Granularity Balancing (AGB) that distinguishes the contributions of representations at different granularities to predictions based on temporal semantics. Additionally, we design a specific temporal loss function to optimize the embeddings of time based on the similarity constraints of adjacent timestamps. Extensive evaluations conducted on four event-based benchmark datasets demonstrate the superiority and effectiveness of $\mathsf{LGRe}$. The contributions of this paper can be summarized as follows:
\begin{itemize}[leftmargin=*]
    \item We propose $\mathsf{LGRe}$, a simple yet effective TKG completion method that sufficiently discovers representations at different granularities of TKGs.
    \item Through time encoding and adaptive granularity balancing, $\mathsf{LGRe}$ effectively allocates predictive contributions from various temporal granularities.
    \item Experiments on four widely-used TKG benchmarks demonstrate that $\mathsf{LGRe}$ outperforms state-of-the-art TKG completion baselines, showcasing its superiority in learning time-related representations.
\end{itemize}
\section{Related Work}
KG completion task, also known as link prediction, has been widely studied to infer missing facts based on existing knowledge. This task is divided into completions on static KGs and TKGs due to the dynamics of knowledge.
\subsection{Static KG Completion}
Static KG completion models focus on representing correlations between entities and relations within triples. Translation-based approaches, such as TransE \cite{transe}, view objects as the result of transformations of subjects and relations in euclidean embedding space. Additionally, methods such as semantic matching \cite{distmult}, tensor decomposition \cite{complex}, and complex space embedding \cite{quatE,rotate} offer different perspectives on analyzing latent interactions within triples. Neural network models, such as ConvE \cite{conve} and CompGCN \cite{compgcn}, demonstrate the effectiveness of Convolutional Neural Networks (CNNs) and Graph Neural Networks (GNNs) in learning interactions among entities and relations, or among neighboring facts. However, existing static methods still fall short in exploiting temporal information and capturing the evolutionary patterns of facts.
\subsection{TKG Completion}
In the realm of transformation, TTransE \cite{ttranse} is the first to propose the temporal ordering relation assumption (i.e., $r_i\textbf{T} \approx r_j$), introducing time into the traditional translation model TransE. ChronoR \cite{chronor} extends RotatE by incorporating time-parametrized rotation transformations. TA-DistMult \cite{tadistmult} adapts DistMult to account for temporal information in learning relation types. HyTE \cite{hyte}, driven by the semantics of time, constructs temporally aware hyperplanes for each timestamp to represent the entity-relation space. DE-SimplE \cite{desimple} introduces a diachronic function to represent entity embeddings at any temporal point. TComplEx \cite{tntcomplex}, BDME \cite{bdme}, Joint-MTComplEx \cite{jmt} and CEC-BD \cite{cec} employ tensor decomposition to involve temporal embeddings in quadruple embeddings. Building on this, TeLM \cite{telm} treats each relation as a dual multivector to enhance generalization in TKG representation. BoxTE \cite{boxte} introduces unique temporal information for each relation using box embeddings. Recently, SANe \cite{SANe} incorporates a time-aware parameter generator to capture the dynamic features of facts using CNNs, and TGeomE++ \cite{TGeo} leverages a geometric algebra-based product to model diverse entities, relations, and temporal dynamics. However, many facts exhibit different semantics at various temporal levels. Most existing methods treat the semantics of timestamps as holistic, independent vectors, neglecting to learn representations at different temporal granularities for each fact. This approach overlooks the analysis of how these varied granularities influence fact representation and reasoning.
\section{Method}
\begin{figure*}[b!]
\centerline{\includegraphics[width=0.7\linewidth]{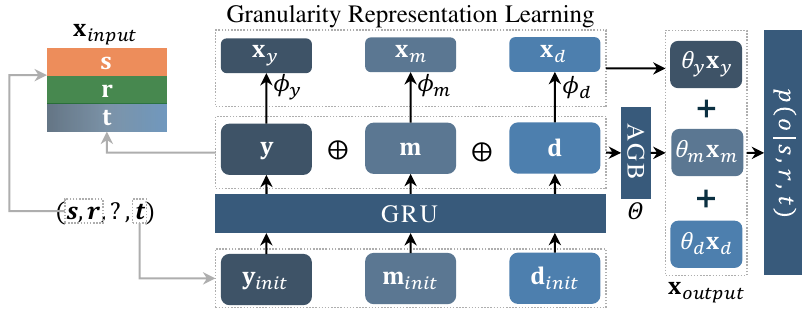}}
\caption{An illustration of $\mathsf{LGRe}$ architecture, where $\mathbf{x}$ denotes the representation of query. $\mathbf{y}$, $\mathbf{m}$, and $\mathbf{d}$ refer to embeddings from different granularities, respectively.}
\label{framework}
\end{figure*}
The framework of $\mathsf{LGRe}$ is shown in Figure \ref{framework}. It maintains a simple encoder-decoder architecture containing two key modules: (a) Granularity Representation Learning (\S \ref{GRL}), which captures interactions among entities, relations, and timestamps through multi-layer convolutional operators and generates representations at different granularities, and (b) Adaptive Granularity Balancing (\S \ref{AGB}), which assigns different weights to representations based on embeddings of temporal semantics. Notably, the decoder only necessitates a simple linear transformation since the above modules effectively fuse the features of quadruples.
\subsection{Notations}
In this paper, Temporal Knowledge Graphs (TKGs) are represented as a set of quadruples with temporal information, denoted as $\mathcal{G}=\{(s,r,o,t) | s, o \in \mathcal{E}, r \in \mathcal{R}, t \in \mathcal{T}\}$. Here, $\mathcal{E}$, $\mathcal{R}$, and $\mathcal{T}$ represent the sets of entities, relations, and timestamps, respectively. Each quadruple $(s,r,o,t)$ in $\mathcal{G}$ is distributed in a specific timestamp $t$. Given a query set $\mathcal{Q}=\{(s,r,?,t)|s \in \mathcal{E}, r \in \mathcal{R}, t \in \mathcal{T}\}$, the objective of TKG completion is to predict the missing object $o$ at some known timestamps, which is also considered as a link prediction task. In this paper, lowercase letters indicate corresponding semantics, and bold letters denote vectors. Table \ref{notations} summarizes the important notations. Notably, while we generally use object prediction as an example, both object and subject predictions are considered in our study.
\begin{table}[ht]
\caption{The Summary of important notations.}
\centering
\begin{tabular}{ll}
\hline
Notations & Descriptions \\ 
\hline
$\mathcal{G}$ & temporal knowledge graph with a set of quadruples\\
$\mathcal{E}$, $\mathcal{R}$, $\mathcal{T}, \mathcal{Q}$  & entity, relation, timestamp, and query set of TKGs\\
$s$, $r$, $t$ & entity, relation and time of a quadruple \\
$\textbf{s}, \textbf{r}, \textbf{t}$  & entity, relation and time embeddings\\
$y$, $m$, $d$ & separate temporal semantics of \textit{year}, \textit{month} and \textit{day} \\
$\textbf{y}$, $\textbf{m}$, $\textbf{d}$ & separate temporal embeddings of \textit{year}, \textit{month} and \textit{day} \\
$\phi$ & the convolutional operator for each layer\\
$\lambda$ & the parameters of convolutional operators \\
$\theta_y,\theta_m,\theta_d$ & adaptive weight of different granularities \\
$\alpha$ & hyper-parameter for temporal regularization coefficient \\
\hline
\end{tabular}
\label{notations}
\end{table}
\subsection{Granularity Representation Learning} \label{GRL}
In practice, temporal information can represent potential periodic semantics (e.g., the same \textit{month} each \textit{year} or the same \textit{day} each \textit{month}). However, representing temporal information as a whole unit fails to capture these patterns effectively. Hence, we decompose time into the format "\textit{year}-\textit{month}-\textit{day}", with each part denoting distinct semantics to preserve the integrity of the temporal information. The separated temporal embeddings are then combined into a sequence format and processed through a Gated Recurrent Unit (GRU) as follows:
\begin{equation}
\{\textbf{y},\textbf{m},\textbf{d}\} = \mathsf{GRU}(\{\textbf{y}_{init},\textbf{m}_{init},\textbf{d}_{init}\}) \textnormal{,}
\end{equation}
where $\{\textbf{y}_{init},\textbf{m}_{init},\textbf{d}_{init}\}$ are initial temporal embeddings from Xavier initialization. Then we combined all temporal embeddings with initial subject $\textbf{s}_{init}$ and relation $\textbf{r}_{init}$ to obtain the initial input of our CNNs:
\begin{equation} \label{RU}
\begin{split}
     &\mathbf{t} = \sigma(\textbf{W}_t[\textbf{y}||\textbf{m}||\textbf{d}]+\textbf{b}_t) \textnormal{,} \\
     &\mathbf{r} = \sigma(\textbf{W}_r[\textbf{r}_{init}||\mathbf{t}]+\textbf{b}_r) \textnormal{,} \\
     &\textbf{x}_{input} = [\textbf{s}_{init}||\mathbf{r}] \textnormal{,} 
\end{split}
\end{equation}
where $\sigma$ denotes \textit{LeakyReLU} activation function, $\textbf{W}_t \in \mathbb{R}^{d \times 3d}$ and $\textbf{W}_r \in \mathbb{R}^{d \times 2d}$ denote linear transformations, $\textbf{b}_t$ and $\textbf{b}_r$ are bias for temporal and relation embeddings.

Similar to SANe \cite{SANe}, we employ a parameter generation mechanism $g(\cdot)$ to determine the parameters of convolutional operators (e.g., $\lambda_y$). Specifically, $\mathsf{LGRe}$ contains three-layer convolutional neural networks, where parameters in each convolutional layer are generated through the linear transformation of the corresponding temporal embeddings (i.e., $\{\textbf{y}, \textbf{m}, \textbf{d}\}$). The specialized parameters are more effective in capturing the dynamic patterns between entities and relations by considering temporal information. Using "\textit{year}" as an example, $\lambda_y$ is generated by $g(y)$, where $\lambda_y \in \mathbb{R}^{c_o \times c_i \times k \times k}$, with $c_o$, $c_i$, and $k$ denoting the output channels, input channels, and kernel size of the convolutional operator, respectively. The embedding of each granularity can then be calculated as follows:
\begin{equation}
\begin{split}
    &\textbf{x}_1 = \phi(\textbf{x}_{input}, \lambda_y) \textnormal{,} \\
    &\textbf{x}_y = \textbf{W}_{\textbf{x}_y} \textbf{x}_1 + \textbf{b}_{\textbf{x}_y} \textnormal{,}
\end{split}
\end{equation}
where $\phi(\cdot)$ denotes each convolutional layer with filter parameters $\lambda_y$ generated by corresponding temporal embedding, $x_y$ represents the embedding of \textit{year}. The dimension of $\textbf{W}_{\textbf{x}_y}$ and $\textbf{b}_{\textbf{x}_y}$ are corresponding to the filter and feature map. $\textbf{x}_1$ is then iteratively fed into the next layer $\phi(\cdot)$ to obtain the subsequent intermediate variables $\textbf{x}_2$, $\textbf{x}_3$ and the final embeddings $\textbf{x}_m$ and $\textbf{x}_d$.
\subsection{Adaptive Granularity Balancing} \label{AGB}
As $\mathsf{LGRe}$ learns granularity representations from the above modules, they should be assigned different weights according to the temporal semantics to form the final representations of queries. The separate representations of time for each query are first transformed into a one-dimensional vector to indicate the weight of the corresponding granularities. The adaptive weight $\{\theta_y,\theta_m,\theta_d\}$ of different granularities can be calculated as follows:
\begin{equation}
    \{\theta_y,\theta_m,\theta_d\} = \sigma [\textbf{W}_y \textbf{y}||\textbf{W}_m \textbf{m}||\textbf{W}_d \textbf{d}] \textnormal{,}
\end{equation}
where $\textbf{W}_y, \textbf{W}_m, \textbf{W}_d \in \mathbb{R}^{1 \times d}$ denote the linear transformations for different granularities, $\sigma$ denotes \textit{softmax} activation function.

The final output of our model can be calculated as follows:
\begin{equation}
    \textbf{x}_{output} = \sigma(\theta_y\textbf{x}_y + \theta_m\textbf{x}_m + \theta_d\textbf{x}_d) \textnormal{,}
\end{equation}
where $\sigma$ denotes \textit{ReLU} activation function. The sequential learning in our model in line with the rationality of Residual Connections. The final prediction scores can be represented as:
\begin{equation}
    p(o|s,r,t) = \sigma(\textbf{x}_{output} \textbf{E}^{\top}) \textnormal{,}
\end{equation}
where $\sigma$ is \textit{Sigmoid} function and $\textbf{E}^{\top}$ denotes the transpose matrix of entity embeddings, used to represent the probability of each entity.

\subsection{Training Objective}
The main objective of training in $\mathsf{LGRe}$ is to minimize the negative log-likelihood loss function:
\begin{equation}
    \mathcal{L}_{main} = -\frac{1}{|\mathcal{G}|} \sum_{(s, r, o, t) \in \mathcal{G}} \left( \log p(o|s,r,t) 
+ |\mathcal{N}_{(s,r,t)}| \sum_{\hat{o} \in \mathcal{N}_{(s,r,t)}} \log (1 - p(\hat{o}|s,r,t))\right) \textnormal{,}
\end{equation}
where the loss measures the \textit{binary cross entropy} between true targets and probabilities, simultaneously distancing the representation from negative samples $\hat{o} \in \mathcal{N}_{(s,r,t)}$ corresponding to each query $(s,r,?,t)$.

Given the role of temporal representations in $\mathsf{LGRe}$, it is crucial to include precise and comprehensive semantics in these representations and consistently optimize them. As our temporal embedding $\textbf{t}$ is consist of separate embeddings of \textit{years}, \textit{months} and \textit{days} to ensure the periodic semantics, the integrated representations of adajacent timestamps should be similar. Therefore, we propose a temporal loss function to smooth the temporal embeddings.
\begin{equation} \label{TL}
    \mathcal{L}_{\mathcal{T}} = \frac{1}{|\mathcal{T}| - 1} \sum_{i=1}^{|\mathcal{T}|-1} ||\mathbf{t}_{i+1}^{T} \mathbf{t}_{i}||_p \textnormal{,}
\end{equation}
where $||\cdot||_p$ denote L-2 normalization. The overall loss $\mathcal{L}$ of $\mathsf{LGRe}$ can be defined as:
\begin{equation}
    \mathcal{L} = \mathcal{L}_{main} + \alpha \mathcal{L}_\mathcal{T} \textnormal{,}
\end{equation}
where $\alpha$ is the coefficient of temporal regularization.
\subsection{Complexity analysis}
$\mathsf{LGRe}$ contains two main modules, Granularity Representation Learning (GRL) and Adaptive Granularity Balancing (AGB). We set batch size as $m$ with embedding dimension $d$. The computational complexity of GRL is $\mathcal{O}(3md^2 + \lambda m d)$ since it employs GRU and convolutional operators. The complexity of AGB is $\mathcal{O}(m d^2 + m d)$ as we only use linear transformations and sum operations in AGB. The final prediction with transpose has a complexity of $\mathcal{O}(|\mathcal{E}|md)$. Thus, the overall computational complexity of training can be simplified to $\mathcal{O}(m d^2 + (\lambda + |\mathcal{E}|) m d)$. Additionally, the space complexity of $\mathsf{LGRe}$ is $\mathcal{O}(|\mathcal{E}|d + |\mathcal{R}|d + |\mathcal{T}|d)$ since we simultaneously consider entity, relation, and time encoding in our model.
\section{Experiments}
\subsection{Experimental Setup}
In this section, we provide detailed description of benchmarks, baselines and evaluation metrics employed in our study, as well as the implementation details of the proposed $\mathsf{LGRe}$.
\subsubsection{Datasets.}
To demonstrate the effectiveness and applicability of our method, we utilize four widely used benchmark datasets with different granularities in this study. These include two datasets from the Integrated Crisis Early Warning System (ICEWS): ICEWS14 and ICEWS05-15 \cite{desimple}, and two public KGs: YAGO11k and Wikidata12k \cite{hyte}. The facts in ICEWS14 and ICEWS05-15 are annotated with timestamps in the format "\textit{year}-\textit{month}-\textit{day}", while YAGO11k and Wikidata12k only provide \textit{year} granularity. Following previous works \cite{SANe}, we introduce fabricated timestamps with constant values for months and days. Detailed information about the datasets is shown in Table \ref{dataset}.
\begin{table*}[htbp]
\centering
\caption{Statistics of ICEWS14, ICEWS05-15, YAGO11k and Wikidata12k datasets.}
\begin{tabular}{ccccccccc}
\hline
Dataset & Entities & Relations & Train & Valid & Test & Timestamps & Granularity \\
\hline
ICEWS14 & 7,128 & 230 & 74,845 & 8,514 & 7,371  & 365 & 1 day \\
ICEWS05-15 & 10,488 & 251 & 368,868 & 46,302 & 46,159 & 4,017 & 1 day \\
YAGO11k & 10623 & 10 & 16,408 & 2,050 & 2,051 & 72 & 1 year \\
Wikidata12k & 7,691 & 240 & 1,734,399 & 238,765 & 305,241 & 82 & 1 year \\
\hline
\end{tabular}%
\label{dataset}
\end{table*}

\subsubsection{Baselines.}
The proposed $\mathsf{LGRe}$ is compared with static KG completion methods \cite{transe,distmult,rotate,complex,quatE} and the state-of-the-art TKG completion models, including TTransE \cite{ttranse}, HyTE \cite{hyte}, TA-DistMult \cite{tadistmult}, DE-SimplE \cite{desimple}, TComplEx \cite{tntcomplex}, ChronoR \cite{chronor}, TeLM \cite{telm}, BoxTE \cite{boxte}, SANe \cite{SANe}, TGeomE++ \cite{TGeo}, BDME \cite{bdme}, Joint-MTComplEx \cite{jmt} and CEC-BD \cite{cec}. Compared with static KG completion models, TKG models are more capable to capture temporal dependencies and evolutionary features within TKGs.
\subsubsection{Evaluation Metrics.}
In our experiments, we utilize the widely-used Mean Reciprocal Rank (MRR) and Hits@\{1,3,10\} as evaluation metrics. MRR is computed as the average of the reciprocals of the rank positions of the correct answers for all queries. Hits@\{1,3,10\} measures the accuracy of the top-\{1,3,10\} retrieved entities by determining the proportion of queries for which the correct entity is ranked within the positions of the retrieved list.
\begin{table*}[h]
\centering
\caption{
    Link prediction results on ICEWS14 and ICEWS05-15 datasets. All results are reported from the previous studies. Dashes indicate unobtainable results. The best results are written bold.
    }
    \label{results1}
\begin{tabular}{ccccccccc}
\toprule
\multirow{2}{*}{Model} &\multicolumn{4}{c}{ICEWS14}&\multicolumn{4}{c}{ICEWS05-15}\cr 
 \cmidrule(lr){2-5}\cmidrule(lr){6-9}
&MRR&Hits@1&Hits@3&Hits@10&MRR&Hits@1&Hits@3&Hits@10\cr
\midrule
        TransE &.280&.094&-&.637&.294&.090&-&.663\cr
        DistMult &.439&.323&-&.672&.456&.337&-&.691\cr
        RotatE &.418&.291&.478&.690&.304&.164&.355&.595\cr
        ComplEx &.467&.347&.527&.716&.481&.362&.535&.729\cr
        QuatE &.471&.353&.530&.712&.482&.370&.529&.727\cr
\midrule
        TTransE &.255 &.074 &- &.601 &.271 &.084 &- &.616 \cr
        HyTE  &.297 &.108 &.416 &.655 &.316 &.116 &.445 &.681  \cr
        TA-DistMult &.477 &.363 &- &.686 &.474 &.346 &- &.728 \cr
        DE-SimplE&.526  &.418 &.592  &.725 &.513  &.392  &.578  &.748 \cr
        TComplEx&.610&.530&.660&.770&.660&.590&.710&.800\cr
        ChronoR&.625&.547&.669&.773&.675&.596&.723&.820\cr
        TeLM&.625&.545&.673&.774&.678&.599&.728&.823\cr
        BoxTE&.613&.528&.664&.763&.667&.582&.719&.820\cr
        SANe &.638&.558&.688&.782&.683&.605&.734&.823\cr
        TGeomE++&.629 &.546 &.680 &.780 &.686 &.605 &.736 &.833\cr
        BDME & .635& .555 & .683 & .778 &- &- &- &-\cr
        Joint-MTComplEx & .636 & .556 & .681 & .786 & .683 & .601 & .736 & .832 \cr
        CEC-BD & .633 & .554 & .680 & .777 & .681 & .602 & .730 & .825\cr
\midrule
        $\mathsf{LGRe}$&\textbf{.644} &\textbf{.564} &\textbf{.694} &\textbf{.790} &\textbf{.695} &\textbf{.616} &\textbf{.746} &\textbf{.838}\cr
\bottomrule
\end{tabular}
\end{table*}
\subsubsection{Implementation Details.}
Following previous work \cite{SANe}, the initial embedding dimension $d$ is configured to 300 for ICEWS05-15 and 200 for other datasets. The Adam optimizer is employed with learning rate, batch size and negative sampling number parameterized to 0.001, 256 and 1000, respectively. The dropout rate of 0.2 is applied uniformly across all modules. The convolutional kernel size is set in the range of \{3, 5, 7\}. We conduct grid search to select the optimal hyper-parameter $\alpha$ in the range of \{1e-5, 5e-5, 1e-4, 5e-4, ..., 0.1\}. In details, $\alpha$ is set to 1e-5, 5e-4, 5e-6 and 5e-4 for ICEWS14, ICEWS05-15, YAGO11k and Wikidata12k, respectively. All experiments are conducted on NVIDIA RTX 4090 GPUs. Notably, the raw and inverse query (i.e., $(s,r,?,t)$ and $(o,r^{-1},?,t)$) are both considered in our study.
\subsection{Main Results}
The main results for the four datasets are shown in Table \ref{results1} and Table \ref{results2}. Above the middle horizontal line are completion methods on static KGs, while below are TKG completion methods. We observe that the early studies in TKG completion, which are mostly based on static methods, do not significantly outperform those static models. This indicates that simply introducing or combining time is not sufficient to capture temporal dependencies, as the unique dynamics in TKGs require deeper consideration. On this basis, newer studies that incorporate temporal information into entities and relations provide a more innovative use of time. Consequently, $\mathsf{LGRe}$ outperforms the static methods (e.g., QuatE) with significant average improvements of 57.39\% and 93.69\% on MRR and Hits@1 across all benchmarks, demonstrating the importance of considering temporal information from multiple perspectives. Additionally, compared with the SOTAs of TKG completion, $\mathsf{LGRe}$ still achieves average improvements of 1.83\% and 2.59\% on MRR and Hits@1, showcasing the effectiveness of considering the interactions between entities and relations at different time granularities.

Specifically, compared with these advanced methods on event-based datasets with multi-granularity in Table \ref{results1}, $\mathsf{LGRe}$ excels at learning representations with complex interactions between entities and relations joint with different time granularities. This combination captures the dynamic knowledge within facts and reflects their evolutionary patterns. Furthermore, $\mathsf{LGRe}$ can assign different weights to these representations, as the semantics of the same entity or relation may change across different time granularities. As a result, it exhibits average improvements of 1.13\% and 1.45\% on MRR and Hits@1.
\begin{table*}[t!]
\centering
\caption{
    Link prediction results on YAGO11k and Wikidata12k datasets. The best results are written bold.
    }
    \label{results2}
\begin{tabular}{ccccccccc}
\toprule
\multirow{2}{*}{Model} &\multicolumn{4}{c}{YAGO11k}&\multicolumn{4}{c}{Wikidata12k}\cr 
 \cmidrule(lr){2-5}\cmidrule(lr){6-9}
&MRR&Hits@1&Hits@3&Hits@10&MRR&Hits@1&Hits@3&Hits@10\cr
\midrule
        TransE &.100&.015&.138&.244&.178&.100&.192&.339\cr
        DistMult &.158&.107&.161&.268&.222&.119&.238&.460\cr
        RotatE &.167&.103&.167&.305&.221&.116&.236&.461\cr
        ComplEx &.167&.106&.154&.282&.233&.123&.253&.436\cr
        QuatE&.164&.107&.148&.270&.230&.125&.243&.416\cr
\midrule
        TTransE&.108 &.020 &.150 &.251 &.172 &.096 &.184 &.329 \cr
        HyTE &.105&.015 &.143&.272 &.180&.098&.197&.333 \cr
        TA-DistMult &.161&.103 &.171&.292&.218&.122&.232&.447\cr
        TComplEx&.185&.127&.183&.307&.331&.233&.357&.539\cr
        TeLM&.191&.129&.194&.321&.332&.231&.360&.542\cr
        SANe &.250&.180&.266&.401&.432&.331&.483&.640\cr
        TGeomE++&.195 &.130 &.196 &.326 &.333 &.232 &.362 &.546\cr
        BDME & .223 & .162 & .227 & .350 & .339 & .241 & .371 & .548\cr
        Joint-MTComplEx & .222 & .158 & .229 & .356 & .360 & .270 & .393 & .541 \cr
        CEC-BD & .212 & .154 & .215 & .339 & .339 & .241 & .369 & .543\cr
\midrule
        $\mathsf{LGRe}$&\textbf{.258} &\textbf{.188} &\textbf{.274} &\textbf{.402} &\textbf{.440} &\textbf{.341} &\textbf{.491} &\textbf{.642}\cr
\bottomrule
\end{tabular}
\end{table*}

Moreover, compared with SOTAs on common-sense and pedia KGs (in Table \ref{results2}), which have coarse-grained but large time spans, $\mathsf{LGRe}$ achieves average improvements of 2.53\% and 3.73\% on MRR and Hits@1. The results show that $\mathsf{LGRe}$ can smooth the influence of large-grained time (e.g., "year") through pseudo-labels and uncover deeper associations using multi-layer convolutions to optimize the overall representation quality. Additionally, through the temporal loss, $\mathsf{LGRe}$ can model the similarity of timestamps in adjacent years. This allows the model to express similar semantics for recent facts with the combination of entities and relations, aligning with the evolution pattern of facts.
\subsection{Ablation Study}
To demonstrate the effectiveness of our key innovation and proposed modules, we conduct ablation studies on all benchmarks, which are shown in Table \ref{ablation}.
\begin{table*}[htbp]
\centering
\caption{Ablation studies on four datasets.}
\begin{tabular}{ccccccccccccc}
\toprule
\multirow{2}{*}{Variant Model} & \multicolumn{3}{c}{ICEWS14} & \multicolumn{3}{c}{ICEWS05-15} & \multicolumn{3}{c}{YAGO11k} & \multicolumn{3}{c}{Wikidata12k} \\ \cmidrule(lr){2-4} \cmidrule(lr){5-7} \cmidrule(lr){8-10} \cmidrule(lr){11-13} 
& MRR & H@1 & H@3 & MRR & H@1 & H@3 & MRR & H@1 & H@3 & MRR & H@1 & H@3\\ 
\midrule
$\mathsf{LGRe}$-w/o-$\mathsf{RU}$ & .641 & .561 & .689 & .694 & \textbf{.616} & .745 & .255 & .185 & .266 & .436 & .335 & .484 \\
$\mathsf{LGRe}$-w/o-$\mathsf{AGB}$ & .638 & .558 & .688 & .678 & .597 & .728 & .254 & .187 & .265 & .438 & .340 & .490 \\
$\mathsf{LGRe}$-w/o-$\mathsf{TL}$ & .641 & .561 & .691 & .693 & .614 & .744 & .256 & .185 & .271 & .436 & .340 & .485 \\
\midrule
$\mathsf{LGRe}$ &\textbf{.644} &\textbf{.564} &\textbf{.694} &\textbf{.695} &\textbf{.616} &\textbf{.746} &\textbf{.258} &\textbf{.188} &\textbf{.274} &\textbf{.440} &\textbf{.341} &\textbf{.491}\\
\bottomrule
\end{tabular}%
\label{ablation}
\end{table*}
\subsubsection{Effect of $\mathsf{RU}$.} Relation Updating ($\mathsf{RU}$) denotes the linear transformation and combination of our initial input in granularity representation learning, corresponding to equation (\ref{RU}). It is beneficial to capture a comprehensive representation of time from multiple granularities, endowing each fact with dynamism through multiple linear layers combined with relations and entities. The results show that $\mathsf{RU}$ enhances $\mathsf{LGRe}$ with average improvements of 0.68\% and 0.99\% on MRR and Hits@1, demonstrating the importance of representing dynamic knowledge in encoding. Notably, more significant improvements (1.05\% and 1.71\% on MRR and Hits@1) are achieved on YAGO11k and Wikidata12k. Unlike ICEWS, both of these datasets span long time periods, while their facts are long-lived, potentially spanning multiple centuries. Therefore, incorporating temporal information into these facts is crucial to better reflect long-term dynamic semantics.

\subsubsection{Effect of $\mathsf{AGB}$.} Adaptive Granularity Balancing ($\mathsf{AGB}$), which achieves average improvements of 1.37\% and 1.28\% on MRR and Hits@1, is used to assign different weight for representations from different granularities. Specifically, performance of $\mathsf{AGB}$ is more prominent on the multi-granularity or large-scale datasets like ICEWS05-15 (with improvements of 2.51\% and 3.18\% on MRR and Hits@1), which has a "\textit{year}-\textit{month}-\textit{day}" granularity and across eleven years. This fine-grained raw time format highlights the effectiveness of AGB. Additionally, in YAGO11k and Wikidata12k, $\mathsf{AGB}$ can smooth the effect of \textit{year} by constant values of \textit{month} and \textit{day} to obtain better scores. Overall, the results demonstrate the importance of learning representations from different granularities, with varying degrees of attention.
\begin{figure}[t!]
\centering
    \subfloat[Results of different $\alpha$ on ICEWS14.]{
        \includegraphics[width=0.4\linewidth]{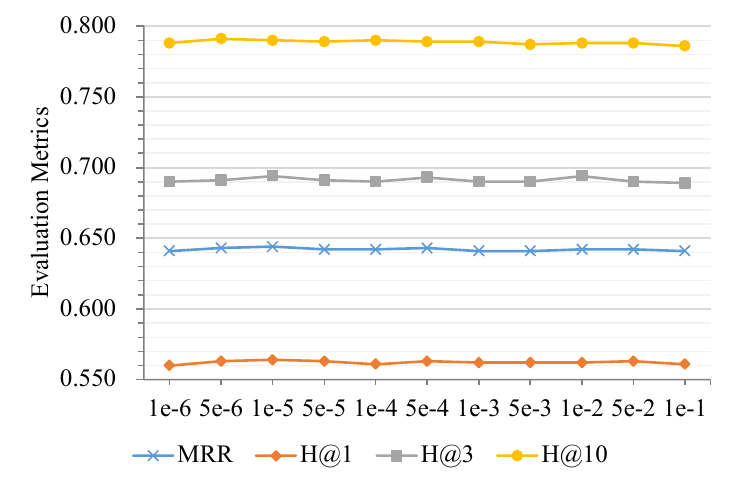}}
        \centering
    \subfloat[Results of different $\alpha$ on ICEWS05-15.]{
        \includegraphics[width=0.4\linewidth]{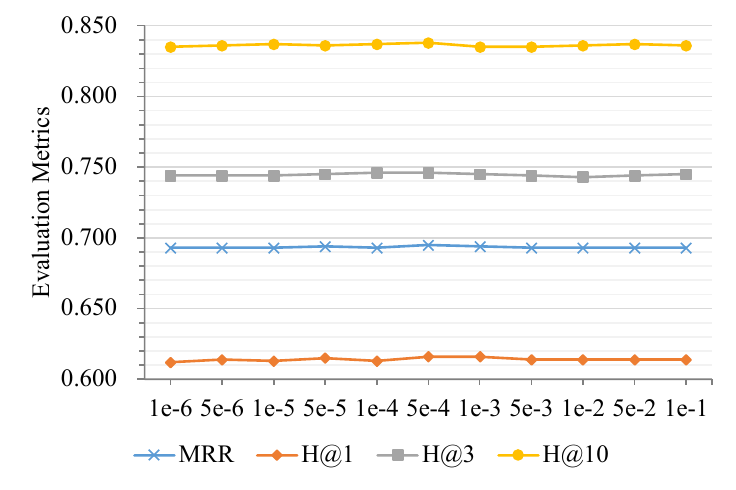}}
        \centering
    \\
    \subfloat[Results of different $\alpha$ on YAGO11k.]{
        \includegraphics[width=0.4\linewidth]{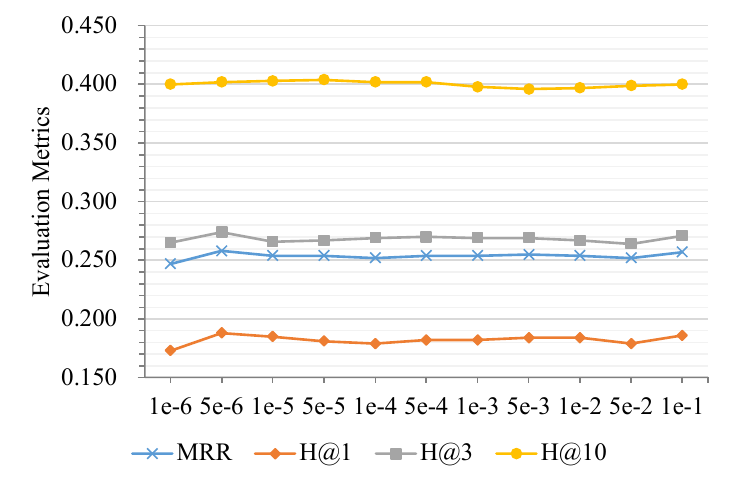}}
        \centering
    \subfloat[Results of different $\alpha$ on Wikidata12k.]{
        \includegraphics[width=0.4\linewidth]{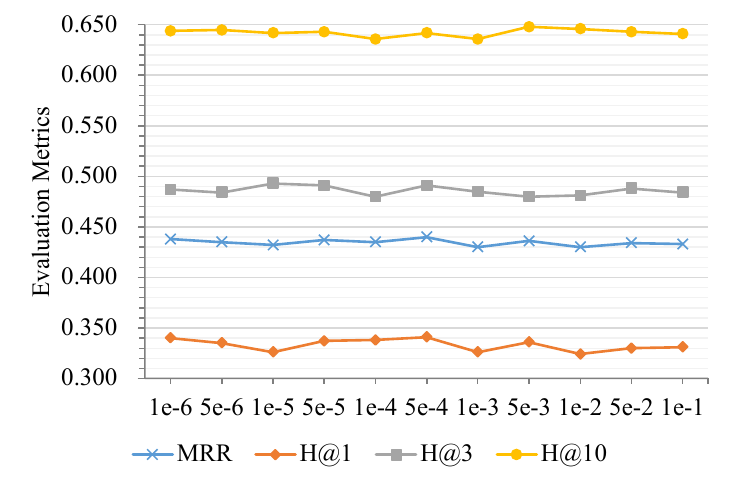}}
        \centering
    \caption{Link prediction results with different $\alpha$ settings on ICEWS14, ICEWS05-15, YAGO11k and Wikidata12k.}
    \label{hyp}
\end{figure}
\subsubsection{Effect of $\mathsf{TL}$.} Temporal Loss ($\mathsf{TL}$) is employed to control the regularization coefficient in temporal representation learning. It ensures that adjacent timestamps have similar semantics, reflecting the similar characteristics of real-time adjacent facts. Consequently, $\mathsf{TL}$ enhances $\mathsf{LGRe}$ by 0.62\% and 0.70\% on MRR and Hits@1, demonstrating its effectiveness.
Notably, $\mathsf{TL}$ shows even greater efficiency on pedia datasets, with gains of 0.85\% and 0.96\% on MRR and Hits@1. We suggest that due to the extremely large time span, the factual associations may expire and are more sensitive to temporal semantics. This observation aligns with the analysis in "Effect of $\mathsf{RU}$".
\subsection{Sensitivity Analysis}
In the sensitivity analysis, we assess the impact of the hyper-parameter $\alpha$, which governs the temporal regularization coefficient to optimize temporal embeddings. The value is determined over a smaller scale range of \{1e-5, 5e-5, 1e-4, 5e-4, ..., 0.1\} to ensure that the main loss works. The results show that $\mathsf{LGRe}$ maintains stable performance across all benchmarks. Specifically, for the ICEWS datasets, which have finer time granularity but smaller time spans, the model is less sensitive to temporal semantic similarity. In contrast, for YAGO11k and Wikidata12k, which have much larger time spans, the model is more sensitive to the learning of temporal semantics. Therefore, better time representation is beneficial to their overall performance.
\subsection{Visualization of Weight Analysis}
\begin{figure*}[htbp]
\centerline{\includegraphics[width=0.9\linewidth]{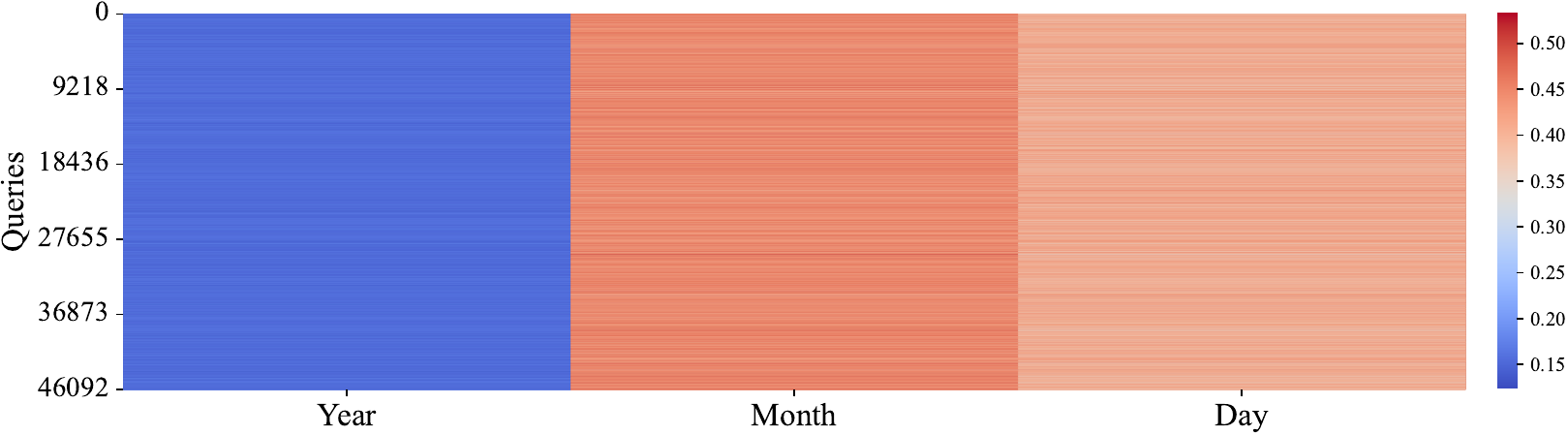}}
\caption{Weight assignment of AGB. The \textit{y}-axis represents each query, and the \textit{x}-axis represents the weights of \textit{Year}, \textit{Month}, and \textit{Day} for the query.}
\label{weight}
\end{figure*}

To investigate the effectiveness of considering multiple granularities for queries, as well as the unique influence from each granularity, we visualize the weight assignment of adaptive granularity balancing in $\mathsf{LGRe}$. We adopt object prediction on ICEWS05-15 as an example, shown in Figure \ref{weight}. The average weights for \textit{year}, \textit{month}, and \textit{day} are 0.1573, 0.4394, and 0.4032, respectively. It can be observed that the weight of the \textit{year} representation is lower than that of \textit{month} and \textit{day} in most cases. Additionally, for the latter two temporal semantics, 72.25\% of queries rely more on \textit{month} representations than \textit{day}, indicating that many facts in real-world maintain similar semantics over longer periods (e.g., \textit{month}). As a consequence, each fact exhibits unequal dependence on the three temporal representations, further demonstrating the importance of considering and distinguishing multiple granularities simultaneously.
\section{Conclusion}
In this paper, we propose $\mathsf{LGRe}$, a simple yet effective method for TKG completion that excels at learning varied TKG representations based on different temporal granularities. It encompasses two key modules: 1) granularity representation learning, which achieves diverse representations of entities and relations at the granularity level according to discrete temporal semantics, and 2) adaptive granularity balancing, which adaptively generates corresponding weights for unique embeddings based on the temporal semantics of each granularity to make the final prediction. Extensive experiments on four datasets with dynamic knowledge demonstrate the superiority and effectiveness of $\mathsf{LGRe}$. Future research directions include investigating more efficient encoding methods for entities and relations, and exploring deeper dynamic interactions between entities.

%
%
%
\bibliographystyle{splncs04}
\bibliography{mypaper}
\end{document}